# Data Association for an Adaptive Multi-target Particle Filter Tracking System

**Raphael Alampay**
Ateneo de Manila University
Quezon City, 1108 Philippines
+632-4266001 ext 5660
raphael.alampay@gmail.com

**Kardi Teknomo**
Ateneo de Manila University
Quezon City, 1108 Philippines
+632-4266001 ext 5660
teknomo@gmail.com

**ABSTRACT**
This paper presents a novel approach to improve the accuracy of tracking multiple objects in a static scene using a particle filter system by introducing a data association step, a state queue for the collection of tracked objects and adaptive parameters to the system. The data association step makes use of the object detection phase and appearance model to determine if the approximated targets given by the particle filter step match the given set of detected objects. The remaining detected objects are used as information to instantiate new objects for tracking. State queues are also used for each tracked object to deal with occlusion events and occlusion recovery. Finally we present how the parameters adjust to occlusion events. The adaptive property of the system is also used for possible occlusion recovery. Results of the system are then compared to a ground truth data set for performance evaluation. Our system produced accurate results and was able to handle partially occluded objects as well as proper occlusion recovery from tracking multiple objects.

## Introduction

Video tracking is defined as the process of estimating the location of one or more objects in a video or from a camera or video file [4]. The objects being tracked by the system depends on the application it addresses. It may be defined as any moving object in the scene, pedestrians or cars. The collection of state information of these objects for every time step is referred to as an object's trajectory.

The trajectory information brought by a video tracking system can be used for many applications. Surveillance is one of the most popular applications for video tracking systems such as the one used by IBM to monitor activities of objects within the scene of a compound [2]. Other examples include monitoring pedestrian activity and classifying their trajectories to complex behaviors as done by [6] and creating probabilistic models for pedestrian flow from captured trajectories [5].

All applications of a video tracking system have to deal with several common problems. The first problem is noise, which is defined as unwanted or false information brought about by the vision sensor [4]. The amount of noise often depends on the quality of the sensor being used. Another major problem is occlusion wherein targets are either failed to be observed or tracked properly if obscured by other valid targets or foreign objects in the scene (i.e. target moves behind a wall or two or more objects merging together) [4].

In order to deal with the problems of noise and occlusion, multiple methods have been used. Multiple object tracking by [11] was done by using probability trees which takes into account the tracking configuration of several previous frames on its histogram values, distance and speed. Using a robust likelihood model, the system is able to track objects and deal with minor occlusions. Another approach by [13] used an observation model to determine occlusions. In this approach, occlusion is detected by locating significant decreases in similarity values when comparing a tracked object to a reference object.

One of the recently popular methods for tracking is the use of particle filter which was first introduced by Isard and Blake [3]. The appealing reasons to use the particle filter method for tracking multiple objects are mainly its ability to deal with non-linear state space and its multi-modal property both of which contribute to the ability of the particle filter to deal with partial occlusions. Particle filter is based on Monte Carlo sampling. The samples it produce, when matched with some observation model, can create non-Gaussian and/or nonlinear scores. In the context of tracking, this approach is useful to overcome clutter between objects or when an object has some parts of it covered by other objects. With the particle filter approach, the system will be able to model the partially available information from samples which give a higher chance for the tracker to track the objects even in the case of partial occlusion. More details regarding the particle filter will be discussed in the later sections.

Though several papers (e.g. see [9] for survey) have shown tracking multiple objects using particle filter successfully, there are several important problems that they did not address. One of the main problems is *target hijacking*. Target hijacking occurs when two or more tracked objects merge with each other causing occlusion and at the same time causes the particle filter to update its current state to the wrong object. Other studies have dealt with this problem using learned approaches such as modifying the motion model of the system to take into account previous number of velocity data of the tracked object and adding it as information to compute the most probable state in the succeeding frames [7]. The problem with such approaches is that it assumes the object being tracked moves linearly and may not cope with sudden change of direction.

Our approach takes advantage of the multi-modal capabilities of the particle filter to track through partial occlusion and at the same time introduce a data association step coupled with object detection and an appearance model to deal with target hijacking. Data association in a traditional sense is used to match every object in every scene to compute one's likelihood that it is indeed part of previous objects. It is often a needed step for track-before-detect systems such as those discussed by [10] but often increases computation complexity as the number of objects increase. Our approach reduces the data association step by considering only nearby detected objects per tracked object as opposed to associating all objects in the scene. This however is done prior to particle filter (meaning a separate routine in addition to and after the filter) and as a means of verification therby adding a layer of computation. Throughout the discussion of the paper, we will show how these steps, when coupled with an adaptive parameter approach, deal with noise, occlusion events (partial and full) and occlusion recovery.

The rest of the paper is divided as follows. We first discuss the particle filter approach in detail and present some examples that have modified this approach to deal with tracking problems. We then present our methodology by first giving an overview and then breaking it down to the different algorithms used by the particle filter, data association and the models used by the system particularly the appearance model and transition model. The last few sections of the paper discuss the system's performance against ground truth data.

**Related Literature**

The particle filter approach is based on the sequential Monte Carlo method. Random numbers is utilized to estimate the state of an object in a time series. In the case of video tracking, the state is commonly the x and y coordinates of the object in the 2D image plane (video frame). In order to generate the next probable state of the object, particle filter requires a transition model which mathematically models the movement of an object. These state representations are the "particles" and are then weighted according to some appearance model. The highest weighted particle is most likely the estimated state of the object. The process is then repeated and updated recursively for tracking throughout the video [4].

The particle filter in the context of video tracking is defined by [1] as a Monte Carlo Bayesian algorithm that uses particles wherein each particle represents a state of the object of interest. The algorithm is Bayesian because it recursively computes for the approximated state of the object by re-instantiating particles from previous states possibly from the actual or predicted prior. In order to do this, the particle filter requires a transition model that defines the movement of the object. Given an object state in a certain frame, particles are then generated for the next frame which represents possible states of the object.

Each particle is then weighted against the reference object using a likelihood or appearance model. The particle with the highest weight is treated as the approximated state of the object at the given frame. The state of the object is then updated and particles are re-sampled for the next iteration. In weighing particles, some systems will define which particles are considered weak and strong. Re-sampling procedure then generates lost particles from either the strongest weighted particles or set of strong surviving particles.

The following is a mathematical representation of the particle filter as discussed by [4]. For explanation of notations, please refer to the notations section in the methodology section. The densities $p_{k|k}(x_k \mid z_{1:k})$ are approximated with a sum of $L_k$ Dirac $\delta$ functions (the particles) centered in $\{x_k^{(i)}\}_{i=1}^{L_k}$ as

$$p_{k|k}(x_k \mid z_{1:k}) \approx \sum_{i=1}^{L_k} w_k^{(i)} \delta(x_k - x_k^{(i)}) \quad (1)$$

where $\{w_k^{(i)}\}_{i=1}^{L_k}$ are the weights associated with the particles and are defined as

$$w_k^{(i)} \propto \frac{p_{k|k}(x_k^{(i)} \mid z_{1:k})}{q_k(x_k^{(i)} \mid z_{1:k})} \quad i=1,\ldots,L_k \quad (2)$$

q() is the importance density function defined as the density that generated the current set of particles.

Assuming that $p_{k-1|k-1}(x_{k-1} \mid z_{1:k-1})$ is approximated by the set of particles and associated weights given by $\{w_{k-1}^{(i)}, x_{k-1}^{(i)}\}_{i=1}^{L_{k-1}}$, by substituting this approximation we can obtain the following form

$$p_{k|k-1}(x_k \mid z_{1:k-1}) \approx \sum_{j=1}^{L_{k-1}} w_{k-1}^{(j)} f_{k|k-1}(x_k \mid x_{k-1}^{(j)}). \quad (3)$$

Recursively, the formulation to propagate the particles and their corresponding weights can be written as

$$w_k^{(i)} \propto \frac{g_k(z_k \mid x_k^{(i)}) \sum_{j=1}^{L_{k-1}} w_{k-1}^{(j)} f_{k|k-1}(x_k^{(i)} \mid x_{k-1}^{(j)})}{q_k(x_k^{(i)} \mid z_{1:k})} \quad (4)$$

Isard and Blake [3] proposed a form where the particles are drawn from the predicted prior as such:

$$q_k(x_k \mid z_{1:k}) = p_{k|k-1}(x_k \mid z_{1:k-1}) \quad (5)$$

The way particles are propagated in a scene depends on the transition model that represents how an object moves. Ghaeminia et al [1] used a modified ARMA (autoregressive moving average) model for its motion model defining how particles are propagated (and thus sampling how the object moves in the scene). The parameters of their transition model depend on the values of the velocity and acceleration displacement of a previous number of frames. During this training period, a simple mean shift tracking process is applied. Although this may accurately model the way an object may move in real life, it is still prone to fail if occlusion occurs in early parts of the video.

Another adaptation of the particle filter in multi-object tracking is done by Wang et al [13]. Unlike Ghaeminia's approach, the way Wang propagated their particles at every time step was by using a simple random walk in which the movement of the object is defined by adding some random noise due to the uncertainty of motion. Instead of focusing on the transition model, they instead implemented a robust likelihood model that deals with variation of scale and is able to detect occlusion by continuous decrease in its likelihood values.

In the next section, we provide our methodology in applying the particle filter in tracking multiple objects as well as some additions to the process to increase its accuracy and deal with occlusion problems.

## Methodology
*Overview*
Tracking multiple objects in the system requires the original frame at a given index as its initial input. We then perform background subtraction to get the objects to either initialize them for tracking or pass them to the tracking module. Tracking then involves utilizing the particle filter algorithm for each tracked object then passing its output to the data association algorithm to determine trajectory state, occlusion state or initialization of new objects. The system is implemented in C++ and uses the OpenCV library framework.

*Notations and Definitions*
For uniformity in the computations to follow, we shall use the following notations:

- $M$ – Number of particles per tracked object
- $m$ – particle index
- $Y$ – Observations from image data ($y$ will then represent each element)
- $X$ – Approximated state or tracked object configuration ($x$ will then represent each tracked object)
- $Z$ – Observed objects from object detection module ($z$ will then represent each observed object)
- $\pi$ – Likelihood value when computing similarities between histograms with range from 0 to 1.
- $T_l$ – Threshold for likelihood with range from 0 to 1. We used a value of 0.6 for our experiments.
- $L$ – The computed likelihood score of the approximated object state after the particle filter.

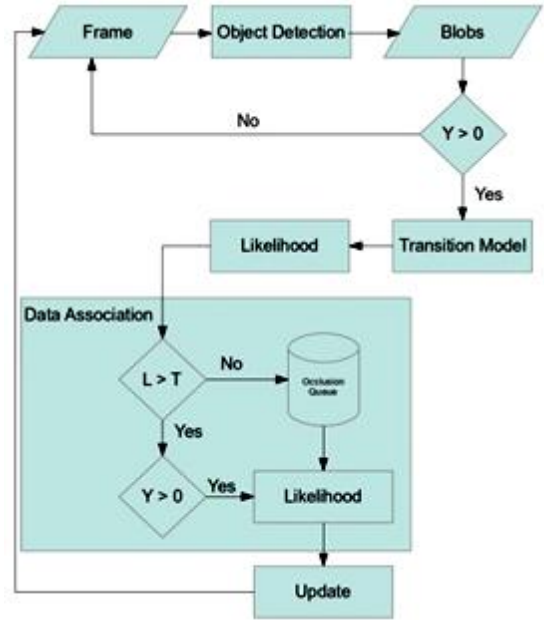

**Figure 1: Overview of system**

The entire system flow is illustrated in Figure 1. The system requires an input of a frame, which is basically a two dimensional array that contains the frame image data as captured by OpenCV. It is stored in a Mat data structure which is a native data type to the OpenCV framework. The system also processes blob data ("Blobs" in the diagram). Blob data contains grouped pixels that represent an object in the scene as detected by the object detection module. It also contains information such as the coordinates of the object's center of mass as well as the rectangular area of the blob's image region (for histogram computation later on). The system also stores an occlusion queue which houses all tracked objects which are in an occluded state but are not dropped for possible recovery. The set of Y is checked if there is at least 1 element present. If there are no elements present, the system doesn't process anything.

*Background Subtraction Algorithm*
For every given frame at index $t$ in the video, the system first performs object detection to define a set of $Z$ objects in the scene. This is done by performing an image processing technique called background subtraction wherein objects in a foreground are extracted as blobs by using a reference background image and getting the difference of pixel values from frame to frame.

The system assumes that the video is taken from a static camera in order to implement a background reference image. The way the reference image is computed is by getting the running average value of each pixel (also called moving average) which involves getting the average value of a temporal signal that takes into account the latest values

received [11]. It can be computed using the following equation:

$$\mu_t = (1-\alpha)\mu_{t-1} + \alpha p_t \tag{6}$$

where $\mu_t$ is the running average of the pixel $p$ at time $t$ and α is a parameter called a learning rate that defines the current value over the currently estimated average. A lower α would mean a faster adaptation to changes in the observed value as suggested by [12]. For the experiment a value of 0.01 was used for $\alpha$.

One of the problems with background subtraction approach is the inherent noise produced both in the background and the foreground. Noise here can be produced if the background contains moving pixels throughout the video which are not exactly objects being tracked which causes false detection (i.e. swaying trees and shadows).

*Particle Filter Algorithm*
The particle filter algorithm recursively approximates the state of an object being tracked by making use of "particles" that represent possible states at a given frame index. The objective is to use a Markovian assumption to approximate the current hidden states given the set of observations. It can be written as a Bayesian filtering distribution using the following:

$$p(x_t|y_t) \approx p(y_t|x_t)\int_{x_{t-1}} p(x_t|x_{t-1})p(x_{t-1}|y_{1:t-1})dx_{t-1} \tag{7}$$

where $p(x_t|y_t)$ is the current state, $p(y_t|x_t)$ is the observation model, $p(x_t|x_{t-1})$ is the transition model and $p(x_{t-1}|y_{1:t-1})$ is the previous object state.

The current state is the approximated state of the object being tracked which then becomes the previous object state during the next iteration. The observation model represents the likelihood function or how we measure the likelihood of the object being in that specific state [7]. In terms of computer vision, the observation model will be represented by comparing histograms of image regions and will be discussed in the next section. The transition model specifies how objects move from frame to frame and is used to propagate the particles.

The steps in performing a single iteration of the particle filter can be then summarized in four steps namely predict measure, update and resample as suggested by [7]. The update step is where the data association phase will take place and will be discussed in the next section. A single iteration of the particle filter is given by the following pseudo code taken from [7]:

```
// START
For each X as x_t
 Current particle set: {x_t^m}m=1...M
// Prediction step
 For m=1...M
// Transition model
x_{t+1}'^m ←transition(x_t^m)
// Compute histogram of the region given by
the transitioned particle
x_{q:t+1}'^m ←computeHisto(x_{t+1}'^m, x_t^m)
 End
// Measurement step
 For m=1...M
// Assign weights to the particles
x_{w:t+1}'^m ←compareHisto(x_{q:t}^m, x_{q:t+1}'^m)
// Normalize particles
Normalize({x_{t+1}'^m}m=1...M)
 End
// Select the most likely particle
x_{q:t+1}'^{a:m} ←MAX({x_{t+1}'^m}m=1...M)
// Update
x_{t+1} ← x_{t+1}'^{a:m}
// Resample
{x_{t+1}^m}m=1...M ←resample({x_{t+1}^m}m=1...M)
End
// END
```

Note that $x_t$ will not always be $x_{t+1}'^{a:m}$ due to the data association step. For normalization and selection process, we use the same algorithm as explained in [7] in which the particle with the highest likelihood will be considered the approximated state. The next algorithm will use data association to refine the likelihood and detect occlusion events, occlusion recovery without instantiate new objects.

*Data Association Algorithm*
In the data association step, the system takes advantage of the observation module to validate the tracked objects. Unlike previously discussed particle filter tracking systems such as that in [10], wherein object detection is only done at the start to define the objects to be tracked, our system uses object detection at every frame step and validates the likelihood of the tracked object $x_{t+1}$ if a detected object falls within its region (bounding rectangular area defined by the scale of the object). If the likelihood is high enough, then we associate the observed object as the tracked object. Else, we determine if the likelihood of $x_{t+1}$ given by $x_{t+1}'^{a:m}$ is high enough. If it is, then it is considered to be tracked. Else, we consider it to be occluded (low value for highest approximated particle) and push it to the occlusion queue. The advantage of this is that if $x_{t+1}'^{a:m}$ indeed has a high enough value and object detection fails to detect an object in that area, then tracking can still continue. Once data association takes place, the remaining observed objects are first compared to currently occluded and tracked objects. If it finds a tracked object with a high enough likelihood value, then that object will be considered to be recovered from occlusion state. All other remaining objects will be initialized as new objects to be tracked. The algorithm for data association is given by the following pseudo code:

```
Given X, X', Z
For each X as x_{t+1}:
    For each Z as z_{t+1}
        If isWithinRange(z_{t+1}, x_{t+1})
            z_{q:t+1} ← computeHisto(z_{t+1}, x_{t+1})
            π ← compareHisto(z_{q:t+1}, x_{q:t+1})
        If π >= T_l
            x_{t+1} ← z_{t+1}
            // Remove from Z
            POP(z_{t+1})
        Else if x_{w:t+1} < T_l
            // Push to occlusion queue
            PUSH(x_{t+1})
        End
    End
End
If count(Z) > 0
    For each Z as z_{t+1}
        For each X' as x_{t+1}
            π ← compareHisto(z_{q:t+1}, x_{t+1})
            If π >= T_l
                x_{q:t+1} ← z_{q:t+1}
                // Reset the parameters of x_{t+1}
                Reset(x_{t+1})
            POP(x_{t+1})
        End
    End
End
If count(Z) > 0
    For each Z as z_{t+1}
        // Initialize as new object
        z_{t+1} ← init(z_{t+1})
        PUSH(z_{t+1})
    End
End
```

It is main parts of the data association as shown in the pseudo code is looping through all X elements, checking if it is within range from Z elements and computing their likelihood scores through the likelihood model. After all elements have been accounted for, the system can then determine if there are new objects in the scene.

*Transition Model*

The transition model makes use of the second order autoregressive moving average (ARMA) equation. When applied to particles, the system makes use of each particle's parameters as parameters to the equation which is given by the following:

$$x_{t+1} = As_t + Bs_{t-1} + Cw_t \qquad (8)$$

where A, B and C corresponds to the 2$^{nd}$ order ARMA parameters and $s_t$, $s_{t-1}$ refers to vectors corresponding to the difference between the current (t) and original and the difference between the previous (t) and original for the x, y and s (scale) values of the tracked object in order to compute for its (probable) state in t+1. Similar to [7], we use the values 2.0, -1.0 and 1.0 for A, B and C respectively. The value of $w$ is generated randomly with a standard deviation of 1.0 and 0.5 for x and y respectively.

The noise parameters of the model will depend on the values stored in each tracked object. This is where adaptation takes place. If the status of the tracked object is occluded, then the standard deviations for both x and y will increase by a certain threshold which affects the value of the noise in the equation. The parameters will continue to increase while the object is in its occluded state for every index. A counter is also implemented to track the number of continuous increase of parameters of the tracked object before it is dropped. 1.0 was used as increment values for both x and y standard deviations.

*Likelihood Model*

Determining the likelihood of two image regions involves comparing them based on a histogram model. In order to compute the histogram of a given region in the image (defined by the x and y coordinates as its center of mass and a rectangular bounding box that specifies its scale), we will use an HSV (hue, value and saturation) color model as suggested by Perez et al [7]. The needed values in the HSV color space will be the hue and saturation values. The $N_h N_s$ bins will be populated using pixels with saturation and value larger than specified threshold 0.1 and 0.2 respectively. The bins of the resulting histogram are thus defined as $N = N_h N_s + N_v$ bins.

In order to compute the likelihood between two given histograms, we make use of the Bhattacharrya similarity coefficient given by the following equation:

$$d(q', q(x)) = \sqrt{1 - \sum_{n=1}^{N} \sqrt{q'(n) q_t(n;x)}} \qquad (9)$$

where q' is the reference histogram and q(x) is the histogram of the region given by a certain particle x [7].

*Determining Accuracy*
In order to determine the accuracy of the system compared to ground truth, we took the difference between the NTXY of the system and the ground truth using the following equation:

$$e = \sqrt{(x_g - x_s)^2 + (y_g - y_s)^2} \quad (10)$$

where $e$ is the resulting difference, $(x_g, y_g)$ is the coordinate of the ground truth and $(x_s, y_s)$ is the coordinate of the particle filter system in a given frame. If $e$ is less than a user defined threshold, then the track is considered to be correct. The threshold is used as a difference factor since ground truth data was extracted using the head of the person as opposed to the system which bases its coordinates on the object's center of mass. For the experiments to follow, a threshold of 25 was used. Accuracy is then computed by taking the number of correctly tracked frames over total number of frames when the object is tracked.

**Results and Analysis**
To verify the effectiveness of our approach, we developed the prototype program on top of OpenCV library framework in C++.

For the validation of the program, we used two data sets. The first dataset was taken from a pedestrian experiment in Indonesia. This video dataset consist of a set of three people walking across a scene in which towards the end of the video, two of the three people merge together then separate. The trajectories taken by the system are then compared with trajectories taken by manual tracking. Trajectory data is recorded by using the NTXY format where N is the object identifier, T is the frame number and X and Y are the x and y image coordinates of the object. In a second data set, we run our system against a PETS [8] video where two people meet, stall for a moment then split. The following image is the output of the tracking system. It displays the trajectory of each tracked object using a different color for each object.

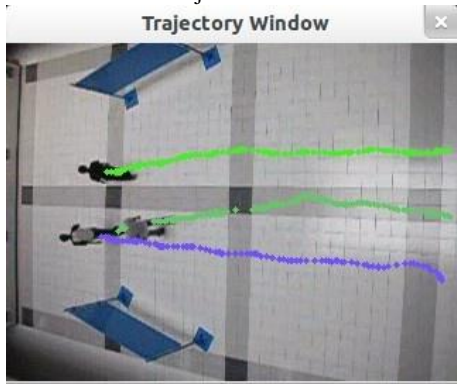

**Figure 2: Occlusion detection and adapting parameters. Even through partial occlusion, the system continues tracking.**

Although the system can accurately track multiple objects even through partial occlusion by using particle filter and data association, the downside of this is computation complexity resulting in a longer processing time compared to other systems mentioned where tracking time is claimed to be able to handle real-time processing. Based on the algorithms in the previous sections, the system is set to run at least $O(mn^2)$. It is expected that as the number of objects being tracked increases, the time it takes for it to compute the trajectories would increase exponentially as well.

The following figure is generated using a Matlab program and shows a statistical representation of the harvested trajectories from the NTXY output of the system in comparison with the NTXY data of the manual system (ground truth):

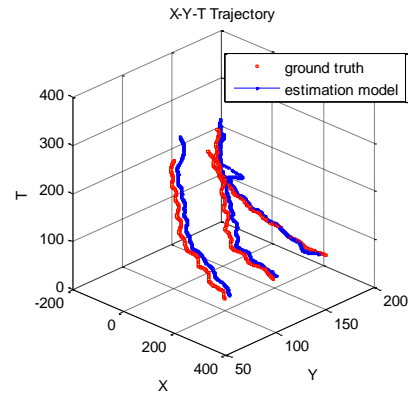

**Figure 3: Trajectory comparisons between ground truth and model estimated.**

For this first data set, occlusion occurs towards the end of the video and is classified as a merging occlusion where one target merges with another. As shown in figure 2, one trajectory steers sideways and is far from the ground truth data. One possible reason for this is that when the two objects merge, the object detection part of the system treats it as one large blob. When passed to the data association part, the HSV values of are compared and is associated with the closer of the two objects. Since one object will be considered to be the most likely one, it will be associated with the detected blob (considering the values are above the threshold) and therefore updating its x and y coordinates to it. Recovery from this occlusion is shown afterwards where one object is recovered and is tracked until the video ends while the other one is failed to be tracked due to low values for its particles even after its parameters have changed. It is also not associated from the data association part of the system since the object was failed to be recognized during those time frames mainly due to a minimum scale threshold of blobs after background subtraction.

The second data set was taken from a kind of standard surveillance data of PETS [8] which exhibited the problem of target hijacking wherein two targets, one coming from the north and one from the south, merge in the middle, stay in that position for a while and split. During the first run,

the system failed to properly track these two objects as shown in the following figure:

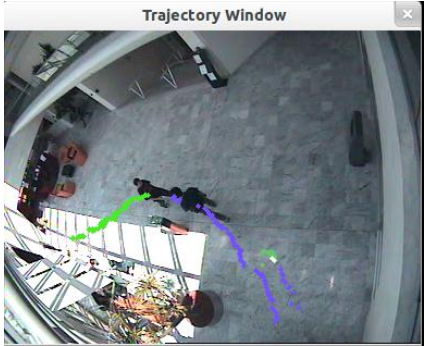

**Figure 4: Before merging**

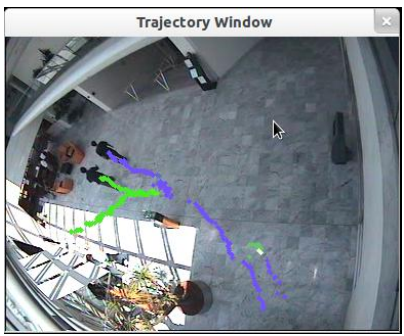

**Figure 5: After splitting, trajectory of objects interchange**

In this case, double target hijacking occurs. One object is updated to the other and continues to follow that object. A set of runs were conducted this time increasing the number of particles per tracked object from 80 to 110. The results turned out to be more accurate but increased the time it took to track the objects. Using 90 or more particles per tracked object produced better and more accurate results as shown by the following figures:

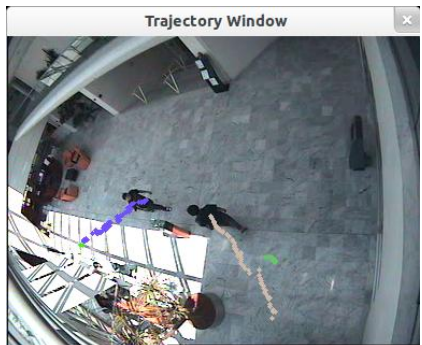

**Figure 6: Before merging**

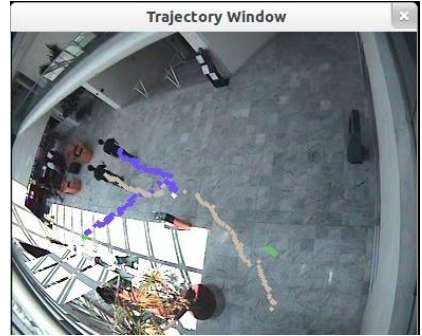

**Figure 7: After split with 90 particles producing correct results**

Objectively, the results are summarized according to time of computation and according to accuracy. Figure 8 shows the computation time against the number of particles.

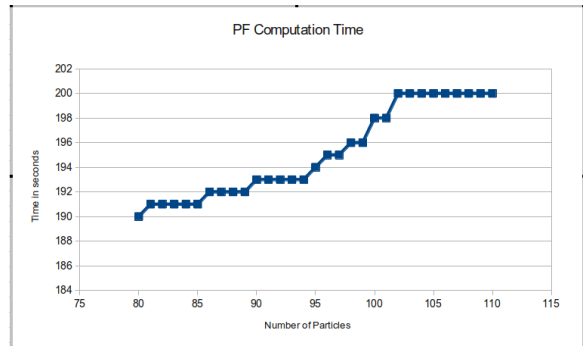

**Figure 8: Computation time in seconds**

Based on these numerical experiments, the computational time increases as the number of particles used increases. For accuracy, we used the error rate for each experiment by taking the inverse of its accuracy (100% minus accuracy percentage).

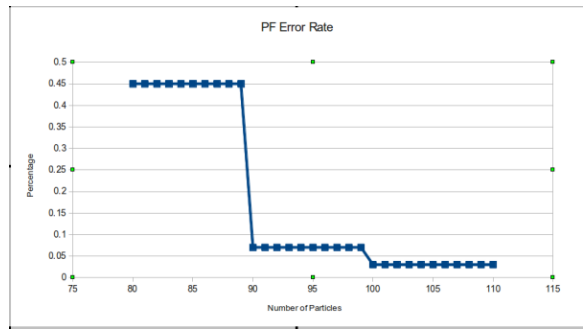

**Figure 9: Error rate for number of particles used**

These results show that using less than 90 particles prevents the system from being able to recover proper tracking after occlusion resulting in target hijacking and a high error rate. 90 or more particles results in accuracy with less than 10% error rate. At the 100 particle mark, accuracy improves.

These suggest that using 100 or more particles would be the optimal choice in order to avoid target hijacking. But after considering both time and accuracy (assuming both have equal weights), we took its change in accuracy, its change in computation time and its cumulative scores and took its ratio. Doing this, we achieved the following results:

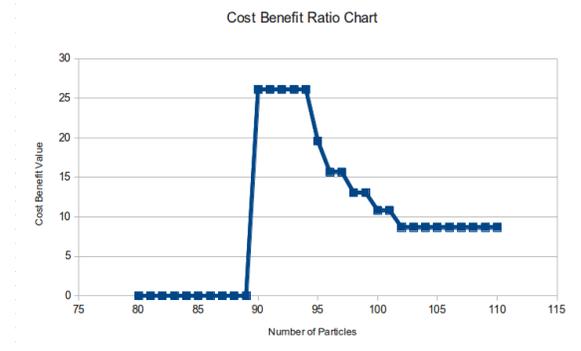

**Figure 10: Cost-Benefit ratio when considering computation time and accuracy**

This shows that the number of particles to be used for optimal performance in terms of time and accuracy to avoid target hijacking is 90 to 95.

**Conclusion and Recommendations**
We have proposed a novel approach to improving the use of particle filter in tracking multiple objects by adding a data association step and keeping track of occlusion states using an occlusion queue. The parameters of each object adjusts depending on its state affecting the way the particle filter "searches" for the object's state throughout each frame and possibly recover it from occlusion. We then utilized the HSV color space of the image for likelihood computation using Battacharyya distance metric. The results are accurate as we are able to track objects within occlusion and recover after a while. However, we sacrificed computation time for adding a data association step. The problem of target hijacking still poses as a challenge for particle filter tracking systems even with the added data association step. Increasing the number of particles increases the system's accuracy but also increases computation time. We have identified an optimal range for number of particles to be used that is able to avoid target hijacking and be as efficient as possible in terms of computation time. It is therefore recommended to find ways to reduce computation time while maintaining the system's accuracy. A better motion model that also follows the Markov assumption is also recommended to better model object movements such as pedestrians. Through our approach, we were able to produce more accurate results by verifying tracking output with data association. Through this verification step, we were also able to handle common problems in video tracking namely occlusion while maintaining proper trajectory for the objects being tracked. By allowing the system to adaptively change some of its parameters, we are able to handle recovery from occlusion.

Our system also proves to perform well with some scenes with possible target hijacking where objects' trajectories tend to interchange by increasing the particle filter's parameters thereby increasing the likelihood of being able to properly track diverging objects from previously merged state.